\documentclass[journal]{IEEEtran}
\usepackage{amsmath,amsfonts}
\usepackage[linesnumbered,ruled,vlined]{algorithm2e}
\usepackage{array}
\usepackage[caption=false,font=normalsize,labelfont=sf,textfont=sf]{subfig}
\usepackage{textcomp}
\usepackage{stfloats}
\usepackage{url}
\usepackage{verbatim}
\usepackage{graphicx}
\usepackage{cite}

\usepackage{graphicx}
\usepackage{textcomp}
\usepackage{adjustbox}
\usepackage{xcolor}
\usepackage{booktabs}
\usepackage{multirow}
\usepackage[inkscapelatex=false]{svg}

\usepackage{amssymb}

\usepackage{amsmath}   
\usepackage{cleveref}  

\newcommand{\rbl}[1]{\textcolor{black}{#1}}

\begin{document}

\title{Tag-Enriched Multi-Attention with Large Language Models for Cross-Domain Sequential Recommendation}

\author{
Wangyu Wu, Xuhang Chen, Zhenhong Chen, Jing-En Jiang, Kim-Fung Tsang,\\ Xiaowei Huang, Fei Ma, Jimin Xiao

\thanks{Jimin Xiao and Fei Ma are corresponding authors.
}
\thanks{
Wangyu Wu is with the Xi'an Jiaotong-Liverpool University, Suzhou, China and the University of Liverpool, Liverpool, UK (email: wangyu.wu@liverpool.ac.uk).}
\thanks{Xuhang Chen is with the School of Computer Science and Engineering, Huizhou University, Huizhou 516007, China (e-mail: xuhangc@hzu.edu.cn).}
\thanks{Zhenhong Chen is with the Microsoft, Redmond, USA (e-mail: zcheh@microsoft.com).}
\thanks{Jing-En Jiang is with the Shandong University, Shandong, China (e-mail: xiaowuga@gmail.com).}
\thanks{Kim-Fung Tsang is with the Shenzhen Institutes of Advanced Technology, Chinese Academy of Sciences, Shenzhen 518055, China (e-mail: kftsang@ieee.org).}
\thanks{Xiaowei Huang is with the University of Liverpool, Liverpool, UK (e-mail: xiaowei.huang@liverpool.ac.uk).}
\thanks{Fei Ma and Jimin Xiao are with the Xi'an Jiaotong-Liverpool University, Suzhou, China (e-mail: fei.ma@xjtlu.edu.cn; jimin.xiao@xjtlu.edu.cn).}
}

\markboth{Journal of \LaTeX\ Class Files,~Vol.~14, No.~8, August~2021}%
{Shell \MakeLowercase{\textit{et al.}}: A Sample Article Using IEEEtran.cls for IEEE Journals}

\IEEEpubid{0000--0000/00\$00.00~\copyright~2021 IEEE}

\maketitle

\begin{abstract}
Cross-Domain Sequential Recommendation (CDSR) plays a crucial role in modern consumer electronics and e-commerce platforms, where users interact with diverse services such as books, movies, and online retail products. These systems must accurately capture both domain-specific and cross-domain behavioral patterns to provide personalized and seamless consumer experiences. To address this challenge, we propose \textbf{TEMA-LLM} (\textit{Tag-Enriched Multi-Attention with Large Language Models}), a practical and effective framework that integrates \textit{Large Language Models (LLMs)} for semantic tag generation and enrichment. Specifically, TEMA-LLM employs LLMs to assign domain-aware prompts and generate descriptive tags from item titles and descriptions. The resulting tag embeddings are fused with item identifiers as well as textual and visual features to construct enhanced item representations. A \textit{Tag-Enriched Multi-Attention} mechanism is then introduced to jointly model user preferences within and across domains, enabling the system to capture complex and evolving consumer interests. Extensive experiments on four large-scale e-commerce datasets demonstrate that TEMA-LLM consistently outperforms state-of-the-art baselines, underscoring the benefits of LLM-based semantic tagging and multi-attention integration for consumer-facing recommendation systems. The proposed approach highlights the potential of LLMs to advance intelligent, user-centric services in the field of consumer electronics.
\end{abstract}


\section{Introduction}
\label{sec:intro}

\begin{figure}[t]
\centering
\includegraphics[width=0.9\linewidth]{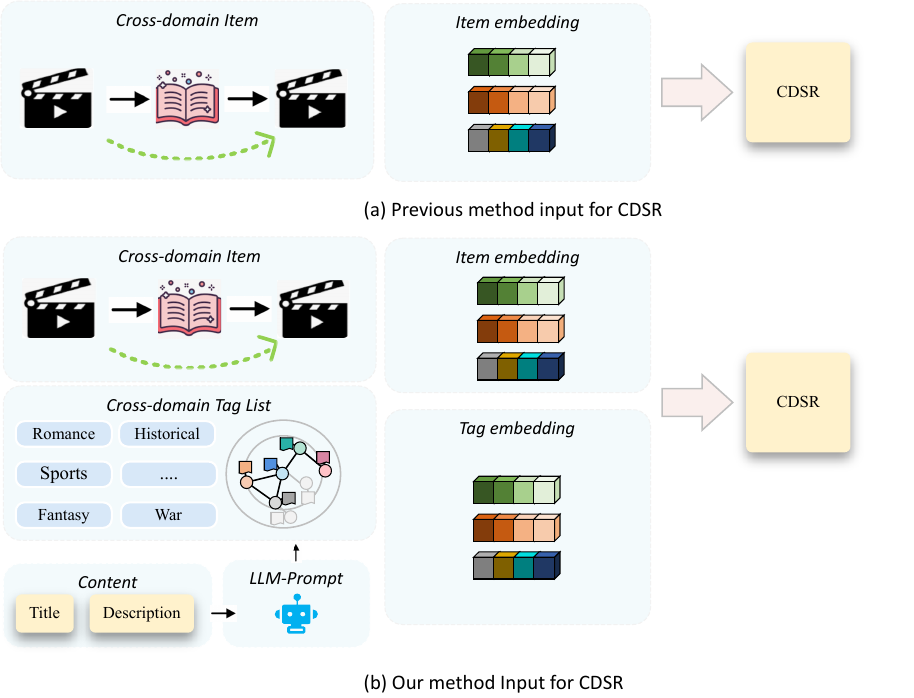}

\caption{(a) Traditional CDSR models primarily rely on item basic features. (b) Our proposed TEMA-LLM enriches item representations by generating semantic tags from LLMs and integrating them with textual and visual features.}
\label{fig:idea}
\end{figure}

\IEEEPARstart{S}{equential} Recommendation (SR) \rbl{has become a widely adopted paradigm for capturing dynamic user preferences by modeling historical interaction sequences~\cite{wang2024autosr,wu2025image}}. It plays a crucial role in recommendation systems~\cite{ahmed2024federated,kong2025oh, Jiang2025revisiting, fathima2025empowering,xue2025structcoh,long2025bridging,long2024coinclip,long2023dynamic,tang2025integralbench,yang2025towards
,zhang2024ai,yu2025bidirectionally,niu2025decoding,zheng2025machine}. While SR methods have achieved strong performance in single-domain scenarios, they remain constrained by issues such as data sparsity and domain-specific bias, where limited interaction records and overfitting to intra-domain patterns hinder generalization. Cross-Domain Sequential Recommendation (CDSR) has emerged as a promising solution by exploiting user behaviors across multiple domains, thereby facilitating knowledge transfer and enabling a more comprehensive view of user interests. This is particularly valuable in modern consumer electronics ecosystems, where users interact with diverse digital services (e.g., shopping, media, fitness) across interconnected smart devices such as smartphones, tablets, wearables, and smart home appliances~\cite{tang2025ocrt,tang2025dissecting}. Despite these advances, current CDSR techniques still face three major challenges: (\emph{i}) they emphasize intra-domain dependencies but often underexplore inter-domain interactions; (\emph{ii}) they insufficiently leverage multimodal information such as item descriptions and images; and (\emph{iii}) even recent LLM-based approaches~\cite{wei2024llmrec} rarely address domain imbalance or generate fine-grained semantic tags for cross-domain alignment. The contrast between previous CDSR and our approach is illustrated in Fig.~\ref{fig:idea}.

\IEEEpubidadjcol
To overcome these limitations, we propose \textbf{TEMA-LLM} (\textit{Tag-Enriched Multi-Attention with Large Language Models}), a novel framework that introduces LLM-driven tag enrichment and a multi-attention mechanism for cross-domain sequential recommendation. Unlike prior LLM-based methods such as LLMRec~\cite{wei2024llmrec}, which primarily augment textual features, TEMA-LLM utilizes LLMs to generate domain-aware prompts and semantic tags from item titles and descriptions, providing richer item semantics and improved cross-domain alignment. Compared with IFCDSR~\cite{wu2025image}, which fuses image and ID embeddings but lacks explicit textual enrichment, our framework enhances item representations by integrating ID, visual, and LLM-generated tag embeddings. Furthermore, in contrast to MAN~\cite{lin2024mixed}, which applies mixed attention without explicitly balancing contributions across domains, TEMA-LLM introduces a tag-enriched multi-attention mechanism that jointly learns intra-domain and inter-domain preferences, thereby capturing complex and evolving consumer interests. To the best of our knowledge,  TEMA-LLM is the first CDSR framework that systematically incorporates LLM-generated semantic tags into a tag-enriched multi-attention design, enabling robust multimodal and cross-domain preference modeling.

\textbf{Our main contributions can be summarized as follows:}
\begin{itemize}
    \item We propose a novel \emph{LLM-based tag enrichment strategy} that generates domain-aware semantic tags from item titles and descriptions, improving semantic alignment and mitigating domain bias in CDSR.
    \item We design a \emph{tag-enriched multi-attention mechanism} that jointly models intra-domain and cross-domain preferences by integrating ID, textual, visual, and tag embeddings, effectively capturing complex user interests. 
    \item We conduct extensive experiments on four large-scale e-commerce datasets, and the results show that TEMA-LLM consistently outperforms recent state-of-the-art baselines~\cite{wu2025image,lin2024mixed}, highlighting the effectiveness of LLM-driven semantic tagging for cross-domain recommendation.
\end{itemize}

\section{Related Work}

\subsection{Sequential Recommendation}

Sequential recommendation~\cite{SRs,liang2025graphical,cai2024relation,wu2025llm,wu2025cognitive} focuses on modeling user interaction histories as ordered sequences, aiming to predict the next likely item conditioned on previously engaged content~\cite{li2024towards,li2024distinct,li2024towardsv2}. \rbl{Traditional methods leveraged first-order Markov chains to capture short-term transitions in user behavior.} As user preferences became more dynamic and behavior patterns more complex, neural network architectures—such as recurrent networks~\cite{GRU, LSTM}, convolutional models~\cite{CNN}, and attention-based structures~\cite{vaswani2017attention}—were incorporated into recommendation frameworks~\cite{DIEN, DIN} to better capture long-range dependencies and evolving interests. More recent developments, including DFAR~\cite{lin2023dual} and DCN~\cite{lin2022dual}, utilize dual-attention architectures to model more sophisticated dependencies and intricate item interactions within sequential data.

Building on these foundations, our approach integrates cross-domain modeling and utilizes large language models (LLMs) to capture complementary semantic signals, boosting generalization and adaptability across diverse content domains.

\subsection{Cross-Domain Sequential Recommendation}

\rbl{In contrast to conventional single-domain sequential recommendation, which captures user behavior confined to a single content space, cross-domain recommendation systems~\cite{xiao2025cross,CDR} alleviate issues such as data sparsity and cold-start by transferring knowledge across domains.}

\rbl{Recent years have witnessed several advances in CDSR. MAN~\cite{lin2024mixed} introduces a mixed attention framework with local and global attention to capture both domain-specific and cross-domain sequential patterns. C2DSR~\cite{cao2022contrastive} further strengthens cross-domain user representations by jointly modeling intra- and inter-sequence item relations with a mutual information objective. More recently, IFCDSR~\cite{wu2025image} explores image fusion strategies to enhance recommendation by integrating visual signals into sequential modeling.}

\rbl{Nonetheless, most existing methods still underexplore explicit multimodal fusion and domain-aware reasoning. Our method addresses these limitations by integrating LLM-generated tag semantics and employing tag-enriched multi-attention to jointly model visual, textual, and ID-based information across domains.}

\subsection{LLM-based Recommendation System}

The integration of Large Language Models (LLMs) into recommender systems has shown promising performance improvements, especially in modeling rich item-level textual features~\cite{wu2023survey,xue2024question,xue2023dual,liang2019adaptive,wang2025not}. \rbl{Representative works include P5~\cite{geng2022recommendation}, which reformulates diverse recommendation tasks into a unified language modeling paradigm, and TALLRec~\cite{bao2023tallrec}, which proposes an efficient tuning framework to align LLMs with recommendation objectives.} While some recent works explore LLMs in cross-domain scenarios~\cite{tang2023one}, they often overlook the joint modeling of textual enhancement, multimodal fusion, and domain-aware reasoning.

In contrast, our method introduces a prompt-based tag generation module to enrich textual semantics across domains, addressing the domain imbalance overlooked by LLMRec~\cite{wei2024llmrec}. Compared with IFCDSR~\cite{wu2025image}, which lacks textual augmentation, our model unifies image, text, and ID representations. Additionally, unlike MAN~\cite{lin2024mixed}, we propose a tag-enriched attention mechanism to explicitly balance domain contributions during fusion.

\begin{figure*}[t] 
\begin{center}
   \includegraphics[width=0.9\linewidth]{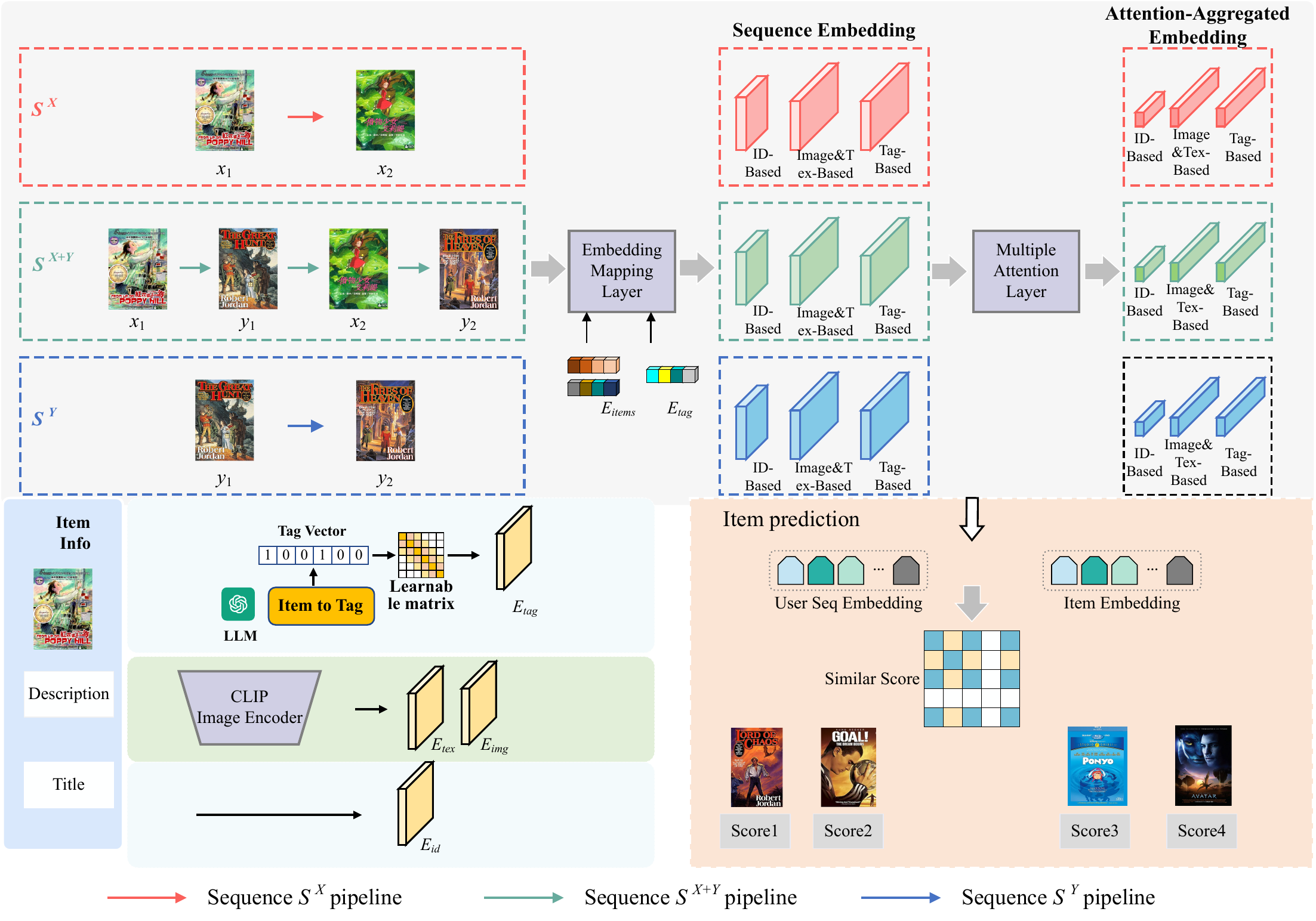}
\vspace{-0.4cm}
\caption{
Overview of the proposed TEMA-LLM framework. 
The \textit{Feature Preparation} module generates multimodal item embeddings using a learnable ID matrix, a frozen CLIP image encoder, and a CLIP text encoder applied to LLM-augmented text. 
Enriched tag embeddings are fused into the item representation via weighted multi-hot encoding. 
These representations are processed by multi-head attention layers to model intra- and inter-sequence user preferences. 
Finally, cosine similarity with candidate embeddings is used for next-item prediction.
}
    \label{fig:framework}
\end{center}
\vspace{-0.5cm}
\end{figure*}

\section{Methodology}
\label{sec:method}

\subsection{Problem Formulation}

In the Cross-Domain Sequential Recommendation (CDSR) setting, users interact with items from two distinct domains, denoted as domain $X$ and domain $Y$, which are associated with item sets $\mathcal{X}$ and $\mathcal{Y}$, respectively. Each user's chronological behavior can be represented as a unified sequence $\mathcal{S}$, composed of three constituent subsequences: $(S^X, S^Y, S^{X+Y}) \in \mathcal{S}$.

Specifically, the domain-specific sequences are defined as 
$$
S^X = [x_1, x_2, \dots, x_{|S^X|}],\ x_i \in \mathcal{X} 
$$
and
$$
S^Y = [y_1, y_2, \dots, y_{|S^Y|}],\ y_i \in \mathcal{Y},
$$
where $|\cdot|$ denotes the sequence length. These capture user interactions that occur solely within domain $X$ and domain $Y$, respectively.

To jointly model cross-domain dynamics, we define the fused sequence $S^{X+Y}$ as the temporal integration of $S^X$ and $S^Y$ into a single ordered list:
\[
S^{X+Y} = [x_1, y_1, x_2, \dots, x_{|S^X|}, \dots, y_{|S^Y|}].
\]
This sequence preserves the original interaction timestamps and enables unified modeling of user preferences across both domains.

The objective of CDSR is to predict the next item the user is most likely to interact with, by estimating the relevance scores or selection probabilities of all candidate items in the combined space $\mathcal{X} \cup \mathcal{Y}$, and selecting the item with the highest predicted likelihood for recommendation.

\subsection{Overall framework}
\label{sec:overall}

\rbl{We propose a unified CDSR framework that leverages LLM-generated tags and multimodal signals to enrich item representations. As shown in Fig.~\ref{fig:framework}, the framework follows a four-stage pipeline, where each module contributes to semantic enhancement, modality fusion, and preference modeling. This design enables effective and interpretable sequential recommendation across heterogeneous domains.}

\rbl{The pipeline starts with a feature preparation module that encodes each item using its ID, image, and LLM-augmented textual inputs. A prompt-driven tag generation module produces semantic tags, which are softly matched to items to form relevance-weighted tag embeddings. These tag features are fused with other modalities into enriched item representations, which are then passed through multi-head attention layers to model intra- and inter-domain preferences. Finally, the next-item prediction is made by computing the similarity between user preference vectors and candidate embeddings.}

\begin{figure*}[t] 
\begin{center}
   \includegraphics[width=1.0\linewidth]{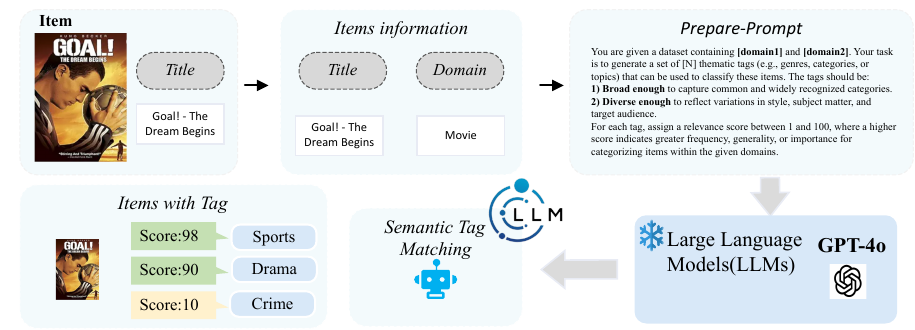}
\vspace{-0.6cm}
\caption{
Overview of the prompt-based enhancement pipeline in our framework.
Given a movie item, we first generate prompts that guide a Large Language Model (LLM) to produce semantic tag information. 
Subsequently, the same LLM is used to compute matching scores between the item and each tag, enabling the selection of relevant tags for embedding generation. 
This two-stage LLM-based process enriches item representations with structured semantic knowledge for downstream recommendation.
}

\label{fig:prompt}
\end{center}
\vspace{-0.5cm}
\end{figure*}

\subsection{Tag List Generation via LLM}
\label{sec:taggen}

To construct a unified tag space for cross-domain sequential recommendation, we employ a self-refine generation strategy that queries a Large Language Model (LLM) multiple times with fixed prompts to generate domain-specific tag lists. This process ensures stability, semantic richness, and consistency across domains.

Given a target domain $d$ (e.g., books or movies), we first define a fixed prompt template $\mathcal{P}_{\text{domain}}$ that instructs the LLM to output $N$ semantic tags relevant to the domain. To address the inherent randomness of LLM decoding, we query the model $R$ times under deterministic settings ($T{=}0$) to obtain multiple candidate tag lists:
\begin{equation}
    \mathcal{T}_d^{(r)} = \text{LLM}(\mathcal{P}_{\text{domain}} \parallel d), \quad r = 1, 2, \dots, R.
\end{equation}
Each response $\mathcal{T}_d^{(r)}$ contains $N$ tag-score pairs, i.e., $\mathcal{T}_d^{(r)} = \{(t_i, s_i)\}_{i=1}^N$, where $t_i$ is a textual tag and $s_i \in [0,1]$ is its LLM-assigned relevance score. We then perform frequency-based majority voting to extract the most representative tag set $\mathcal{T}_d^*$:
\begin{equation}
    \mathcal{T}_d^* = \texttt{Vote}\left( \{\mathcal{T}_d^{(r)}\}_{r=1}^R \right).
\end{equation}
The resulting tag lists $\mathcal{T}_d^*$ across domains form a shared tag vocabulary $\mathcal{T}$ used in downstream modeling. The prompt template is shown in Fig.~\ref{fig:prompt}. The full algorithm for tag list generation is provided in Alg.~\ref{alg:tagselfrefine}.

\begin{algorithm}[t]
\caption{Self-Refine Domain Tag List Generation}
\label{alg:tagselfrefine}
\SetKwInOut{Input}{Input}
\SetKwInOut{Output}{Output}

\Input{Domain identifier $d$, LLM $\mathcal{M}$, prompt template $\mathcal{P}_{\text{domain}}$, number of queries $R$}
\Output{Refined tag list $\mathcal{T}_d^*$}
\BlankLine
\textbf{// Step 1: Multi-query initialization with deterministic decoding} \\
Initialize response set $\mathcal{Z} = \{\}$ \\
\For{$r = 1$ \KwTo $R$}{
    $\mathcal{T}_d^{(r)} = \mathcal{M}(\mathcal{P}_{\text{domain}} \parallel d)$ with $T{=}0$\;
    $\mathcal{Z} = \mathcal{Z} \cup \{\mathcal{T}_d^{(r)}\}$\;
}
\BlankLine
\textbf{// Step 2: Frequency-based voting for stable tags} \\
$\mathcal{T}_d^* = \texttt{Vote}(\mathcal{Z})$\;
\Return $\mathcal{T}_d^*$
\end{algorithm}

\subsection{Tag Embedding Generation}
\label{sec:tagembed}
To enrich the semantic representation of each item, we introduce a tag embedding module based on a predefined tag vocabulary $\mathcal{T} = \{t_1, t_2, \dots, t_{|\mathcal{T}|}\}$. For example, in the book domain, the tag vocabulary may include categories such as \textit{“War”}, \textit{“romance”}, \textit{“history”}, and so on. Each tag $t_i$ is associated with a learnable embedding vector, forming a tag embedding matrix $E_{\text{tag}} \in \mathbb{R}^{|\mathcal{T}| \times d_t}$, where $d_t$ denotes the embedding dimension.

Given an item $v$, we apply a pretrained LLM to extract its relevant tags from its title and description. This results in a soft tag vector $\boldsymbol{w}_v \in [0,1]^{|\mathcal{T}|}$, where each weight $w_i$ reflects the semantic relevance between item $v$ and tag $t_i$.

The final tag embedding $\mathbf{e}_{\text{tag}}^{(v)} \in \mathbb{R}^{d_t}$ is computed via weighted average pooling over the tag embedding matrix:
\begin{equation}
    \mathbf{e}_{\text{tag}}^{(v)} = \sum_{i=1}^{|\mathcal{T}|} w_i \cdot E_{\text{tag}}[i].
\end{equation}
This aggregated representation captures multiple semantic aspects of the item and naturally supports multi-label tagging. It is subsequently concatenated with ID, textual, and visual embeddings to form the final item representation used in downstream modeling.

\subsection{Item Embedding Construction}
\label{sec:itemembed}

To comprehensively represent each item for effective cross-domain sequential recommendation, we construct a unified embedding by combining four types of features: item ID, image, text, and LLM-derived tags. Among them, the first three form the basic multimodal backbone, while the tag embedding introduces semantic interpretability and domain alignment. This section elaborates on the preparation and integration of each component.

\textit{1) Item Embedding via ID Features:}  
We first encode each item using a learnable ID embedding matrix:
\begin{equation}
E_{id} \in \mathbb{R}^{(|\mathcal{X}| + |\mathcal{Y}|) \times d},
\end{equation}
where $|\mathcal{X}| + |\mathcal{Y}|$ denotes the total number of items from both domains, and $d$ is the ID embedding dimension. Each row $E_{id}[v]$ serves as the unique identifier embedding $\mathbf{e}_{id}^{(v)}$ for item $v$, capturing structural identity information.

\textit{2) Visual Embedding via Pretrained Encoders:}  
To represent perceptual content, we extract image features using a frozen CLIP-ViT encoder:
\begin{equation}
E_{img} \in \mathbb{R}^{(|\mathcal{X}| + |\mathcal{Y}|) \times e},
\end{equation}
where $e$ is the CLIP image embedding dimension. For item $v$, its image embedding $\mathbf{e}_{img}^{(v)} = E_{img}[v]$ encodes visual cues such as color, layout, or genre-related imagery (e.g., “swords” for war novels or “faces” for romance movies).

\textit{3) Textual Embedding via CLIP Text Encoder:}  
In parallel, we process item titles using the CLIP text encoder to obtain textual semantic representations:
\begin{equation}
E_{tex} \in \mathbb{R}^{(|\mathcal{X}| + |\mathcal{Y}|) \times e}.
\end{equation}
Each item $v$ receives a textual embedding $\mathbf{e}_{tex}^{(v)} = E_{tex}[v]$, capturing semantic meaning from its title and description.

\textit{4) Tag-Based Semantic Embedding:}  
Following the method in Sec.~\ref{sec:tagembed}, we compute the LLM-derived tag embedding $\mathbf{e}_{tag}^{(v)}$ by weighted average pooling over the tag vocabulary:
\begin{equation}
\mathbf{e}_{tag}^{(v)} = \sum_{i=1}^{|\mathcal{T}|} w_i^{(v)} \cdot E_{\text{tag}}[i],
\end{equation}
where $E_{\text{tag}} \in \mathbb{R}^{|\mathcal{T}| \times d_t}$ is the learnable tag embedding matrix and $w_i^{(v)} \in [0,1]$ reflects the LLM-assigned relevance between item $v$ and tag $t_i$. This step encodes multi-label domain semantics, enabling category-aware alignment across domains.

\textit{5) Similarity Computation with Modal-Specific Embeddings:}  
We denote the fused representation of each modality as a sequence vector (e.g., $\mathbf{h}_{img}$ for visual modality), and compute its relevance with all items via cosine similarity:
\begin{equation}
\text{Sim}(\mathbf{h}_{img}, E_{img}) = \frac{\mathbf{h}_{img} \cdot E_{img}^T}{\|\mathbf{h}_{img}\| \cdot \|E_{img}^T\|},
\end{equation}
where a higher score indicates stronger visual preference alignment. Similar metrics can be computed for textual, ID, and tag modalities.

\textit{6) Final Multimodal Fusion:}  
We integrate the four embedding types by concatenation, followed by an MLP for dimensionality alignment:
\begin{equation}
\mathbf{e}_{item}^{(v)} = \text{MLP}\left( \mathbf{e}_{id}^{(v)} \oplus \mathbf{e}_{img}^{(v)} \oplus \mathbf{e}_{tex}^{(v)} \oplus \mathbf{e}_{tag}^{(v)} \right),
\end{equation}
where $\oplus$ denotes vector concatenation. \rbl{Since the four embedding vectors ($\mathbf{e}_{id}^{(v)}$, $\mathbf{e}_{img}^{(v)}$, $\mathbf{e}_{tex}^{(v)}$, and $\mathbf{e}_{tag}^{(v)}$) may have heterogeneous dimensionalities, the concatenated vector resides in a high-dimensional space. The MLP projects this concatenated vector into a unified latent space of dimension $q$. In practice, we implement the MLP as a two-layer feed-forward network with GELU activation, which ensures both dimensional alignment and non-linear feature interaction. The resulting $\mathbf{e}_{item}^{(v)}$ serves as the final unified embedding for item $v$, which is fed into the downstream multi-attention model.}

 \rbl{This design provides a unified representation that captures structural, perceptual, linguistic, and semantic aspects, which supports both intra- and cross-domain modeling.}

\subsection{Multiple Attention Mechanisms}
\label{sec:Attention}

In traditional sequential recommendation methods~\cite{gru4rec,sasrec}, behavioral sequences from domains $X$ and $Y$ are often processed indiscriminately, \rbl{causing the higher-frequency domain to dominate and skew user preference representation}. To address this, we propose a hierarchical attentional framework that separately processes $S^X$, $S^Y$, and $S^{X+Y}$, \rbl{achieving balanced cross-domain modeling}. Our design adopts a \textit{Tag-Enriched Multi-Attention} strategy that integrates LLM-generated tag embeddings with multimodal features including item ID, image, and text.

We apply a self-attention mechanism within each sequence. For each modality, the sequence $\mathcal{S}$ is first encoded into a corresponding embedding sequence (e.g., $F_{id}$, $F_{img}$, $F_{tex}$, and $F_{tag}$). Then, an attention layer with query-key-value ($\mathbf{QKV}$) structure is applied to capture both short-term and long-term dependencies:
\begin{equation} 
\mathrm{Attention}(\mathbf{Q}, \mathbf{K}, \mathbf{V}) =
\mathrm{softmax}\left(\frac{\mathbf{Q}\mathbf{K}^\top}{\sqrt{d_k}}\right)\mathbf{V},
\end{equation}

\rbl{where $\mathbf{Q}$, $\mathbf{K}$, and $\mathbf{V}$ denote the query, key, and value matrices respectively, and $d_k$ is the key dimensionality.}

\rbl{Each attention module operates independently per (modality, domain) pair, allowing disentangled preference modeling. This structure prevents dominant modalities (e.g., text in books, images in movies) or frequent domains from overwhelming others.}

The attention mechanism outputs the final sequence representation $h$ for each modality and domain. \rbl{Specifically, we compute representations for all combinations of three behavior sequences ($S^X$, $S^Y$, $S^{X+Y}$) and four modalities (ID, image, text, tag), resulting in twelve vectors:}
\begin{align}
&h^{X}_{id},\ h^{X}_{img},\ h^{X}_{tex},\ h^{X}_{tag},\ 
  h^{Y}_{id},\ h^{Y}_{img},\ h^{Y}_{tex},\ h^{Y}_{tag},\nonumber \\
&h^{X+Y}_{id},\ h^{X+Y}_{img},\ h^{X+Y}_{tex},\ h^{X+Y}_{tag}.
\end{align}

\rbl{This systematic design ensures that intra-domain preferences (within $X$ or $Y$) and inter-domain dynamics (across $X+Y$) are both explicitly captured through modality-specific attention.}

Each vector reflects the user’s short-term preference based on a specific modality and domain. For example, the tag-based prediction for domain $X$ is:
\begin{equation}
\mathrm{P}^X_{tag}(x_{t+1} \mid \mathcal{S}) = \text{softmax}\left(\text{Sim}(h_{tag}^X, E_{tag}^X)\right),
\end{equation}
where $E_{tag}^X$ is the domain-specific tag embedding matrix.

Similarly, prediction probabilities based on other modalities are computed:
\begin{equation}
\begin{aligned}
\mathrm{P}^X_{id}(x_{t+1} \mid \mathcal{S})   &= \text{softmax}\left(\text{Sim}(h_{id}^X, E_{id}^X)\right), \\
\mathrm{P}^X_{img}(x_{t+1} \mid \mathcal{S}) &= \text{softmax}\left(\text{Sim}(h_{img}^X, E_{img}^X)\right), \\
\mathrm{P}^X_{tex}(x_{t+1} \mid \mathcal{S}) &= \text{softmax}\left(\text{Sim}(h_{tex}^X, E_{tex}^X)\right).
\end{aligned}
\end{equation}

We then fuse all four predictions via a weighted combination, where for brevity we denote $\mathrm{P}^X_{\cdot}(x_{t+1})$ as $\mathrm{P}^X_{\cdot}(x_{t+1}\mid \mathcal{S})$:
\begin{equation}
\begin{aligned}
\mathrm{P}^X(x_{t+1} \mid \mathcal{S}) = 
&\alpha_1 \mathrm{P}^X_{id}(x_{t+1}) + \alpha_2 \mathrm{P}^X_{img}(x_{t+1}) \nonumber \\
+ \alpha_3 \mathrm{P}^X_{tex}(x_{t+1}) 
&+ (1{-}\alpha_1{-}\alpha_2{-}\alpha_3) \mathrm{P}^X_{tag}(x_{t+1}),
\end{aligned}
\end{equation}
where $\alpha_1$, $\alpha_2$, and $\alpha_3$ control the contributions of each backbone modality. The residual weight is assigned to the tag-based prediction.

To optimize the framework, we minimize the negative log-likelihood loss across domains:
\begin{equation}
\mathcal{L}^X = \sum_{x_t \in S^X} -\log \mathrm{P}^X(x_{t+1} \mid \mathcal{S}),
\end{equation}
with corresponding losses $\mathcal{L}^Y$ and $\mathcal{L}^{X+Y}$ defined similarly. The final training objective is:
\begin{equation}
\mathcal{L} = \mathcal{L}^{X} + \lambda_{1} \mathcal{L}^{Y} + \lambda_{2} \mathcal{L}^{X+Y},
\end{equation}
where $\lambda_1$ and $\lambda_2$ are domain-balancing coefficients.

During inference, we compute the final prediction score for item $x_i$ as:
\begin{equation}
\mathrm{P}(x_i|S) = \mathcal{P}^{X}(x_i|S) + \lambda_{1} \mathcal{P}^{Y}(x_i|S) + \lambda_{2} \mathcal{P}^{X+Y}(x_i|S),
\end{equation}
and select the item with the highest score from the target domain:
\begin{equation}
\mathrm{argmax}_{x_i \in \mathcal{X}} \mathrm{P}(x_i|S).
\end{equation}

\rbl{This attention mechanism is thus not purely heuristic but a modular and empirically grounded design that balances domain-specific signals with cross-modal semantics, enabling robust and interpretable CDSR.}

\section{Experiments}
\label{sec:Experiments}
\subsection{Dataset and Evaluation Metrics}

We evaluate our method on the Amazon dataset~\cite{wei2021contrastive}, a widely used benchmark for Cross-Domain Sequential Recommendation (CDSR). Following~\cite{pinet, mifn, wu2025image}, we select two domain pairs: ``Food-Kitchen'' and ``Movie-Book''. Users with fewer than 10 total interactions or fewer than 3 per domain are removed to ensure sufficient sequence context, and low-frequency items are filtered to reduce sparsity. Final dataset statistics are listed in Table~\ref{sec_exp_tab:dataset}. We adopt a chronological split for training, validation, and testing to simulate real-world usage. Evaluation is based on Mean Reciprocal Rank (MRR)~\cite{mrr} and Normalized Discounted Cumulative Gain (NDCG@5,10)~\cite{ndcg}, assessing both ranking precision and multi-item relevance. Results are averaged over all test users across domain-specific and cross-domain settings.

\textbf{Implementation Details.} To ensure fair comparison, we follow the hyperparameter configurations used in prior work~\cite{mifn}. The learnable embedding dimensionality is set to $q=256$, and CLIP-derived image and text embeddings are fixed at $e=512$. We use a mini-batch size of 256, set the fusion weights as $\alpha_1=0.4$, $\alpha_2=0.2$, and $\alpha_3=0.2$, and apply a dropout rate of 0.3 during training. 
\rbl{We performed a grid search by varying $\lambda_{1}$ within [0.2, 0.5] and $\lambda_{2}$ within [0.05, 0.2]. The framework performs robustly across these ranges, with the best validation score achieved at $\lambda_{1}=0.3$ and $\lambda_{2}=0.1$, which are adopted in all experiments.} 
All models are trained for up to 100 epochs on an NVIDIA 4090 GPU, using the Adam optimizer~\cite{adam}. Hyperparameters are selected via grid search on the validation set, and early stopping with a patience of 10 epochs is applied based on validation loss.

\subsection{Performance Comparisons}

We evaluate TEMA-LLM against a set of state-of-the-art (SOTA) baselines on two cross-domain scenarios: ``Food-Kitchen'' and ``Movie-Book''. Results are reported in Table~\ref{tab:foodkitchen} and Table~\ref{tab:moviebook}. \rbl{Across all metrics, TEMA-LLM consistently achieves the best performance, confirming its effectiveness in modeling both intra- and inter-domain preferences.}

In the Food-Kitchen scenario, \rbl{TEMA-LLM yields the highest scores for both domains. For Food, it achieves 9.34\% MRR, 9.14\% NDCG@5, and 10.32\% NDCG@10. For Kitchen, it obtains 5.24\% MRR, 4.95\% NDCG@5, and 5.50\% NDCG@10, surpassing strong baselines such as LLMRec~\cite{wei2024llmrec} and IFCDSR~\cite{wu2025image}. In contrast, earlier methods like $\pi$-Net, GRU4Rec, and SASRec perform notably worse.}

In the Movie-Book scenario, \rbl{TEMA-LLM again sets a new state of the art. For Movie, it records 6.41\% MRR, 5.44\% NDCG@5, and 5.89\% NDCG@10. For Book, it achieves 2.94\% MRR, 2.62\% NDCG@5, and 2.91\% NDCG@10, outperforming all baselines including LLMRec and IFCDSR. These consistent improvements highlight the benefits of tag-enhanced multimodal features and LLM-guided attention.}


\begin{table}[t]
\centering
\caption{Statistics of two CDSR scenarios.}
\begin{tabular}{cccccc}
\toprule
\textbf{Scenarios}  &\textbf{\#Items} &\textbf{\#Train} &\textbf{\#Valid}  &\textbf{\#Test} &\textbf{Avg.length}\\ \midrule
Food  &29,207  &\multirow{2}{*}{34,117} &2,722   &2,747   &\multirow{2}{*}{9.91}     \\ 
Kitchen  &34,886 &\multirow{2}{*}{}  &5,451   &5,659  &\multirow{2}{*}{}     \\
\midrule
Movie  &36,845  &\multirow{2}{*}{58,515} &2,032  &1,978  &\multirow{2}{*}{11.98}    \\ 
Book  &63,937 &\multirow{2}{*}{}  &5,612  &5,730 &\multirow{2}{*}{}   \\ 
\midrule

\end{tabular}
\label{sec_exp_tab:dataset}
\end{table}

\begin{table}[t]
\footnotesize
\centering
\caption{Experimental results (\%) on the Food-Kitchen scenario.}

\label{tab:foodkitchen}
\setlength\tabcolsep{4.5pt} 
\begin{tabular}{lcccccc}
\toprule
\multirow{2}{*}{Model (Food-Kitchen)} &
\multirow{2}{*}{MRR} &\multicolumn{2}{c}{NDCG} & \multirow{2}{*}{MRR} &\multicolumn{2}{c}{NDCG} \\
\cmidrule(r){3-4}\cmidrule(l){6-7} &(Food)& @5 & @10  &(Kitchen)& @5  & @10   \\
\midrule
$\pi$-Net~\cite{pinet}   &   7.68  &   7.32  & 8.13 &
 3.53 &   2.98  &  3.73  \\

GRU4Rec~\cite{gru4rec}   &   5.79  &   5.48   &  6.13 &
 3.06 &  2.55  &  3.10  \\
SASRec~\cite{sasrec}   &   7.30  &   6.90   &  7.79  &
 3.79 &  3.35  &  3.93  \\ 
SR-GNN~\cite{srgnn}   &    7.84  &   7.58  & 8.35   &
  4.01 &   3.47   &  4.13  \\
PSJNet~\cite{PSJnet}   &   8.33  &   8.07  & 8.77 &
4.10 &  3.68  &  4.32  \\
MIFN~\cite{mifn}  & 8.55  &   8.28  & 9.01   &
  4.09 &   3.57   &  4.29  \\
Tri-CDR~\cite{ma2024triple} &  8.35 &  8.18 &   8.89  &  
4.29  &  3.63 &   4.33  \\
MAN~\cite{lin2024mixed}              & 8.65 & 8.42 & 9.41 & 4.78 & 4.47 & 5.05 \\
LLMRec~\cite{wei2024llmrec}        & 9.05 & 8.86 & 10.03 & 5.02 & 4.73 & 5.34 \\
IFCDSR~\cite{wu2025image} &  9.05 &  8.85 &   9.92  &  
4.95  &  4.56 &   5.24  \\
\textbf{Ours TEMA-LLM} &\textbf{9.34} &\textbf{9.14} &\textbf{10.32}
&\textbf{5.24} &\textbf{4.95} &\textbf{5.50}\\
\bottomrule
\end{tabular}
\end{table}

\begin{table}[t]
\footnotesize
\centering
\caption{Experimental results (\%) on the Movie-Book scenario.}

\label{tab:moviebook}
\setlength\tabcolsep{4.5pt} 
\resizebox{\columnwidth}{!}{ 
\begin{tabular}{lcccccc}
\toprule
\multirow{2}{*}{Model (Movie-Book)} &
\multirow{2}{*}{MRR} &\multicolumn{2}{c}{NDCG} & \multirow{2}{*}{MRR} &\multicolumn{2}{c}{NDCG} \\
\cmidrule(r){3-4}\cmidrule(l){6-7} &(Movie)  & @5 & @10  &(Book)& @5  & @10   \\
\midrule
$\pi$-Net~\cite{pinet}   &   4.16  &   3.72  & 4.17 &
2.17 &   1.84   &  2.03  \\
GRU4Rec~\cite{gru4rec}   &  3.83 &   3.14 &  3.73  &
 1.68 &  1.34   &  1.52  \\

SASRec~\cite{sasrec}   &   3.79 &   3.23 &  3.69  &
 1.81 &  1.41   &  1.71  \\

SR-GNN~\cite{srgnn}   &  3.85 &  3.27   &  3.78 &
  1.78 &  1.40   &  1.66  \\

PSJNet~\cite{PSJnet}  &  4.63 &  4.06 &   4.76 & 
 2.44  &  2.07 &   2.35  \\

MIFN~\cite{mifn}  &  5.05 &  4.21 &   5.20  &  
2.51  &  2.12 &   2.31  \\
MAN~\cite{lin2024mixed}              & 5.95 & 5.06 & 5.45 & 2.64 & 2.34 & 2.57 \\
Tri-CDR~\cite{ma2024triple} &  5.15 &  4.62 &   5.05  &  
2.32  &  2.08 &   2.22  \\
LLMRec~\cite{wei2024llmrec}        & 6.21 & 5.29 & 5.71 & 2.80 & 2.48 & 2.74 \\
IFCDSR~\cite{wu2025image}            & 6.08 & 5.02 & 5.86 & 2.75 & 2.37 & 2.65 \\

\textbf{Ours TEMA-LLM} &\textbf{6.41}  &\textbf{5.44} &\textbf{5.89}  
&\textbf{2.94} &\textbf{2.62} &\textbf{2.91} \\
\bottomrule
\end{tabular}

}
\end{table}

\begin{table}[t]
\centering
\caption{Ablation study on core modules of \textbf{TEMA-LLM} on the Movie domain.}
\label{tab:ablation_main}
\begin{adjustbox}{width=0.95\linewidth}
\begin{tabular}{cccccc}
\toprule
Original Framework & Tag Embedding & LLM Matching & Multi-Attn & MRR \\
\midrule
\checkmark &  &  &  & 5.03  \\
\checkmark & \checkmark &  &  & 5.56 \\
\checkmark & \checkmark & \checkmark &  & 5.89  \\
\checkmark & \checkmark & \checkmark & \checkmark & \textbf{6.41} \\
\bottomrule
\end{tabular}
\end{adjustbox}
\end{table}

\subsection{Efficiency Discussion}
\rbl{While our framework integrates LLM-based tags, CLIP embeddings, and multi-attention layers, the majority of the potentially expensive operations are performed offline. Specifically, LLM queries are only used once during domain tag list construction, and CLIP encoders are applied offline to extract visual/textual embeddings, which are cached for reuse. During online inference, the additional cost comes mainly from the multi-attention module. However, its complexity remains $O(n^2 d)$, the same order as standard self-attention, and thus does not introduce prohibitive overhead compared with existing Transformer-based recommenders.}
\subsection{Ablation Study}

To evaluate the individual contributions of each key component in our proposed TEMA-LLM framework, we conduct ablation studies on the Movie domain by incrementally enabling the following modules: \textit{Tag Embedding}, \textit{LLM Matching}, and \textit{Multi-Attention}. The results are summarized in Table~\ref{tab:ablation_main}. We begin with a baseline model that only utilizes the original sequential recommendation architecture without any tag-enhanced features, yielding an MRR of 5.03\%. By introducing a set of fixed tag embeddings derived from static keyword-based assignment (without leveraging LLM-generated tag–item matching), we observe a notable performance gain to 5.56\%. This suggests that even weak semantic priors from tag information can help improve item representation quality. Next, enabling the LLM-based tag-to-item matching mechanism further boosts the MRR to 5.89\%, highlighting the advantage of incorporating contextual and domain-aware semantics generated by large language models. This component allows the model to dynamically associate relevant tags to each item, resulting in richer semantic alignment. Finally, incorporating the proposed multi-attention mechanism yields the highest performance of 6.41\% MRR. This module enhances cross-modal feature fusion and selectively attends to both domain-specific and shared representations across modalities. The consistent improvement across configurations confirms the effectiveness and complementary roles of all components in the TEMA-LLM framework.

\paragraph{Impact of Tag Representation Strategy.} 
To further analyze the design of our tag embedding generation module, we conduct a focused ablation study on the representation strategies of tag vectors. We compare three alternatives: (1) a standard \textit{one-hot} encoding, where each item is associated with a single most-relevant tag; (2) an \textit{unweighted multi-hot} vector, where multiple tags are assigned without considering their relevance scores; and (3) our full method using a \textit{weighted multi-hot} vector, where relevance scores (ranging from 1 to 100) are normalized and used as soft weights during tag embedding fusion. 

As shown in Table~\ref{tab:ablation_tag}, incorporating multiple tags improves performance over the single-tag one-hot setting, and applying proper relevance-based weighting further enhances the model’s ability to capture fine-grained semantics. These results confirm that both tag diversity and soft weighting contribute significantly to the quality of semantic representations in our framework.

\begin{table}[t]
\centering
\caption{Ablation on different tag representation strategies (Movie domain).}
\setlength\tabcolsep{25pt} 
\label{tab:ablation_tag}
\begin{tabular}{lc}
\toprule
Tag Representation Strategy & MRR \\
\midrule
One-hot tag vector (single tag) & 6.13 \\
Unweighted multi-hot (equal weights) & 6.30 \\
\textbf{Weighted multi-hot (ours)} & \textbf{6.41} \\
\bottomrule
\end{tabular}
\end{table}

\paragraph{Impact of Tag Selection Strategy.} 
We further investigate how different strategies for determining the number of tags per item affect overall performance. Specifically, we compare three representative approaches: (1) selecting the top-$R$ tags with the highest LLM relevance scores to ensure consistency and efficiency; (2) retaining all tags whose scores exceed a predefined threshold $\theta$, which enables semantic coverage but introduces variable-length tag vectors; and (3) a hybrid strategy that first selects the top-$R$ tags and then includes any additional tags with relevance scores exceeding $\theta$, up to a maximum limit. As shown in Table~\ref{tab:ablation_tag_strategy}, the \textit{Top-$R$ only} variant provides a strong baseline due to its fixed-length structure and simplicity. However, the \textit{Hybrid} method achieves the best performance by effectively combining the advantages of coverage and representativeness. In contrast, the \textit{Threshold-only} strategy yields less stable results due to inconsistencies in tag vector length across items, which hinder batch-level learning. These findings highlight the importance of balancing tag diversity and structural consistency for robust semantic representation.
\begin{table}[t]
\centering
\caption{Ablation on tag selection strategies (Movie domain). Top-$R=5$, Threshold-$\theta=70$.}
\setlength\tabcolsep{30pt} 
\label{tab:ablation_tag_strategy}
\begin{tabular}{l|c}
\toprule
Tag Selection Strategy & MRR \\
\midrule
Top-$R$ only (R=5) & 6.29 \\
Threshold-$\theta$ only ($\theta$=70) & 6.14 \\
Top-$R$ + Threshold-$\theta$ (Hybrid) & \textbf{6.41} \\
\bottomrule
\end{tabular}
\end{table}

\section{Conclusion}

In this work, we propose TEMA-LLM, a tag-enriched multi-attention framework for cross-domain sequential recommendation. Our approach leverages LLMs to generate semantically meaningful domain-aware tags, enabling structured and interpretable item representations. \rbl{These tags are encoded as weighted multi-hot vectors reflecting both diversity and importance, and are aligned with CLIP-based vision-language embeddings to form rich multimodal representations. A multi-head attention mechanism jointly models intra- and inter-domain user preferences, facilitating effective knowledge transfer. Experiments on two real-world scenarios show that TEMA-LLM consistently outperforms strong baselines. Ablation studies validate the contributions of LLM-enhanced tag matching, weighted tag fusion, and multi-attention design. Nevertheless, the framework relies on LLM-generated tags and fixed hyperparameters, which may introduce domain bias or limit scalability in dynamic environments. Exploring adaptive strategies and more efficient tag generation mechanisms would be valuable future directions.}

 \section{Acknowledgements}
 This work was supported by the National Natural Science Foundation of China (No. 62471405, 62331003, 62301451), Suzhou Basic Research Program (SYG202316) and XJTLU REF-22-01-010, XJTLU AI University Research Centre, Jiangsu Province Engineering Research Centre of Data Science and Cognitive Computation at XJTLU and SIP AI innovation platform (YZCXPT2022103).

\bibliographystyle{IEEEtran}
\bibliography{mybib}


 





\end{document}